\documentclass[sigconf]{acmart}

\usepackage[utf8]{inputenc}
\usepackage{subfig}
\usepackage{booktabs}
\usepackage{pbox}
\graphicspath{ {./images/} }

\setcopyright{iw3c2w3}
\copyrightyear{2020}
\acmYear{2020}
\acmConference[WWW '20 Companion]{Companion Proceedings of the Web Conference 2020}{April 20--24, 2020}{Taipei, Taiwan}
\acmBooktitle{Companion Proceedings of the Web Conference 2020 (WWW '20 Companion), April 20--24, 2020, Taipei, Taiwan}
\acmPrice{}
\acmDOI{10.1145/3366424.3383542}
\acmISBN{978-1-4503-7024-0/20/04}

\begin{document}

\title[VisBERT: Hidden-State Visualizations for Transformers]{VisBERT: Hidden-State Visualizations for Transformers}

\author{Betty van Aken}
\authornote{Both authors contributed equally to this research.}
\email{bvanaken@beuth-hochschule.de}
\affiliation{%
  \institution{Beuth University of Applied Sciences Berlin}
  \streetaddress{Luxembourg street 10}
  \postcode{13353}
}
\author{Benjamin Winter}
\authornotemark[1]
\email{Benjamin.Winter@beuth-hochschule.de}
\affiliation{%
  \institution{Beuth University of Applied Sciences Berlin}
  \streetaddress{Luxembourg street 10}
  \postcode{13353}
}
\author{Alexander Löser}
\email{aloeser@beuth-hochschule.de}
\affiliation{%
  \institution{Beuth University of Applied Sciences Berlin}
  \streetaddress{Luxembourg street 10}
  \postcode{13353}
}
\author{Felix A. Gers}
\email{gers@beuth-hochschule.de}
\affiliation{%
  \institution{Beuth University of Applied Sciences Berlin}
  \streetaddress{Luxembourg street 10}
  \postcode{13353}
}

\begin{abstract}
Explainability and interpretability are two important concepts, the absence of which can and should impede the application of well-performing neural networks to real-world problems. At the same time, they are difficult to incorporate into the large, black-box models that achieve state-of-the-art results in a multitude of NLP tasks. Bidirectional Encoder Representations from Transformers (BERT) is one such black-box model. It has become a staple architecture to solve many different NLP tasks and has inspired a number of related Transformer models. Understanding how these models draw conclusions is crucial for both their improvement and application. We contribute to this challenge by presenting VisBERT, a tool for visualizing the contextual token representations within BERT for the task of (multi-hop) Question Answering. Instead of analyzing attention weights, we focus on the hidden states resulting from each encoder block within the BERT model. This way we can observe how the semantic representations are transformed throughout the layers of the model. VisBERT enables users to get insights about the model's internal state and to explore its inference steps or potential shortcomings. The tool allows us to identify distinct phases in BERT's transformations that are similar to a traditional NLP pipeline and offer insights during failed predictions.
\end{abstract}

\maketitle

\section{Introduction}
Understanding black-box models is an increasingly prominent area of research. While the performance of neural networks has been steadily improving in nearly every domain, our ability to understand how they work, and how they come to the conclusions they draw is only improving slowly. In order for  large neural networks to be confidently deployed in safety-critical applications, features like transparency, interpretability and explainability are paramount.\\

\noindent\textbf{Visualizing Transformer's Internal States}. One such class of black-box models are Transformer models, BERT in particular. These models have become the state-of-the-art for many different NLP tasks in recent months. While their inherent attention mechanisms offer an avenue for explainability, recent research argues that attention in fact is not ideal for these purposes, or should at least not be fully relied upon \cite{jain2019attention}. We take this as motivation to investigate an approach that might add complementary information. Instead of the attention values, we follow our work in \cite{van2019does} and visualize the hidden states between each BERT layer, and with that the token representations, as they are transformed through the network.\\

\noindent\textbf{Question Answering and Beyond}. The VisBERT tool currently focuses on analyzing the downstream task of Question Answering (QA). QA is a complex task that implicitly requires not only basic language knowledge, but also demands traditional upstream tasks like Named Entity Recognition, Coreference Resolution and Relation Extraction. Besides that, the task often requires multiple inference steps, especially in multi-hop scenarios, which allows us to gain further insights about BERT's reasoning process. We use the three public QA datasets SQuAD \cite{squad}, HotpotQA \cite{hotpot} and bAbI  QA \cite{babi} to show the tool's applicability on three diverse QA tasks including multi-hop reasoning cases.\\
Apart from that, the principle of VisBERT can be easily extended to other up- or downstream NLP tasks. We publish the underlying code\footnote{Code available at \url{https://github.com/bvanaken/visbert}.} in order to enable researchers and practitioners to insert their own models or tasks and to analyze them to gain a better understanding of their inference process. This way potential biases or other shortcomings can be detected and possibly be resolved.\\

\begin{figure*}[t!]
  \centering
  \includegraphics[width=1\textwidth]{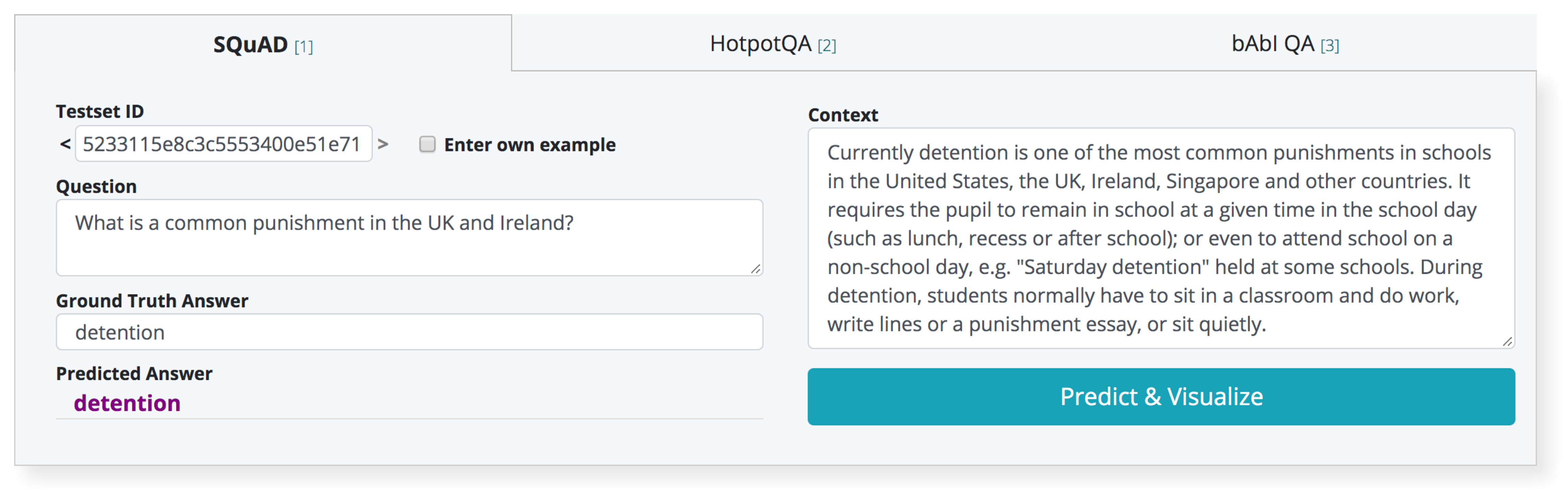}
  \caption{The major control elements of the demo. The top tabs let the user choose between three different fine-tuned BERT models on the QA tasks SQuAD, HotpotQA or bAbI QA. The user can then either choose an example out of the respective test sets, or insert their own example consisting of a context, a question and optionally a ground truth answer. Using the button, the tool presents the predicted answer (in purple) and the visualization of hidden states as shown in Figure \ref{fig:graph}.}
\label{fig:qa-box}
\end{figure*}

\noindent\textbf{Contributions}. The presented work includes the following contributions towards the goal of better understanding Transformer networks:

\begin{itemize}
\item VisBERT\footnote{The tool is available at \url{https://visbert.demo.datexis.com}, a short video demo can be found at \url{https://vimeo.com/383046202}.}, an interactive web tool for interpretable visualization of hidden-states within BERT models fine-tuned on Question Answering.
\item Visualizations of the inference process of unseen examples from three diverse Question Answering datasets, including three BERT (base and large) models fine-tuned on these sets.
\item Identification of four stages of inference that can be observed in all analysed Question Answering tasks.
\item The presented tool allows users to (adversarially) test the abilities and shortcomings of own Question Answering models on arbitrary samples.
\end{itemize}
\section{Visualization of Transformer Representations}
The following section explains the underlying methods used to generate layer-wise visualizations for QA samples in VisBERT.\\

\noindent\textbf{Transformer Models}. 
Transformers \cite{attentionisall} generally consist of three main modules: An embedding module in the beginning, a group of stacked and homogeneous Transformer encoder blocks in the middle, and then either a classification head, or a set of decoder blocks, which mirror the encoder blocks, on top. 
The embedding layer includes a traditional embedding matrix for each token, but Transformers uniquely add a positional embedding as well, in order to introduce a recurrent inductive bias that is not supplied by the attention mechanism. This is in contrast to RNN based networks which inherently contain this recurrent bias.
The classification head for our Question Answering models consists of a single Feed-Forward layer with a Softmax. This head predicts two indices, namely the start and the end index of the answer in the context.
The main representative power of the Transformer lies in its encoder blocks \cite{attentionisall}. Each encoder block includes a multi-headed self-attention module, which transforms each token using the entire input context, normalization, and a Feed-Forward network at the end, which outputs the token representations used by the subsequent layer.\\

\noindent\textbf{Explainability of Transformers}. The architecture of BERT and Transformer networks in general allows us to follow the transformations of each token throughout the network. We use this characteristic for visualizing the changes that are being made to the tokens' representations in each layer. In contrast to analysing the single attention weights within BERT's attention heads as proposed by \cite{attentionvis}, this method allows us to observe the actual outcomes of the whole encoder module in each layer.

Each layer of BERT outputs a different distribution of token vectors and we do not have a reference for semantic meanings of positions within these vector spaces. Therefore we consider distances between token vectors as indication for semantic relations. Following this, we can observe the changing token relations that the model forms throughout the inference process.\\

\noindent\textbf{Processing the Hidden State Representations}.
For a given input QA sample we collect the hidden states from each layer while removing any padding. We then visualize the input on a token-by-token basis. To that end we use the hidden states after each Transformer encoder block, which contains a vector for each token with a dimensionality of 768 (BERT-base) or 1024 (BERT-large). Since these high-dimensional vectors are not directly interpretable we apply dimensionality reduction, mapping the vectors into a two-dimensional space.
As discussed in \cite{van2019does}, among the evaluated dimensionality reduction techniques T-distributed Stochastic Neighbor Embedding (t-SNE) \cite{tsne}, Principal Component Analysis (PCA) \cite{pca} and Independent Component Analysis (ICA) \cite{ica}, PCA is most suitable for this scenario and reveals clusters that correspond to those observed by k-Means clustering \cite{kmeans}. We therefore use PCA for the VisBERT tool and fit it separately for each sample and layer, which allows us to process new samples on the fly. The dimensionality reduction result is a 2D representation of each token throughout the model's layers. We further categorize the tokens based on affiliation to question, supporting facts (facts that are necessary to answer the question) or predicted answer in order to facilitate interpretability.

\begin{figure*}[t!]
  \centering
  \includegraphics[width=0.98\textwidth]{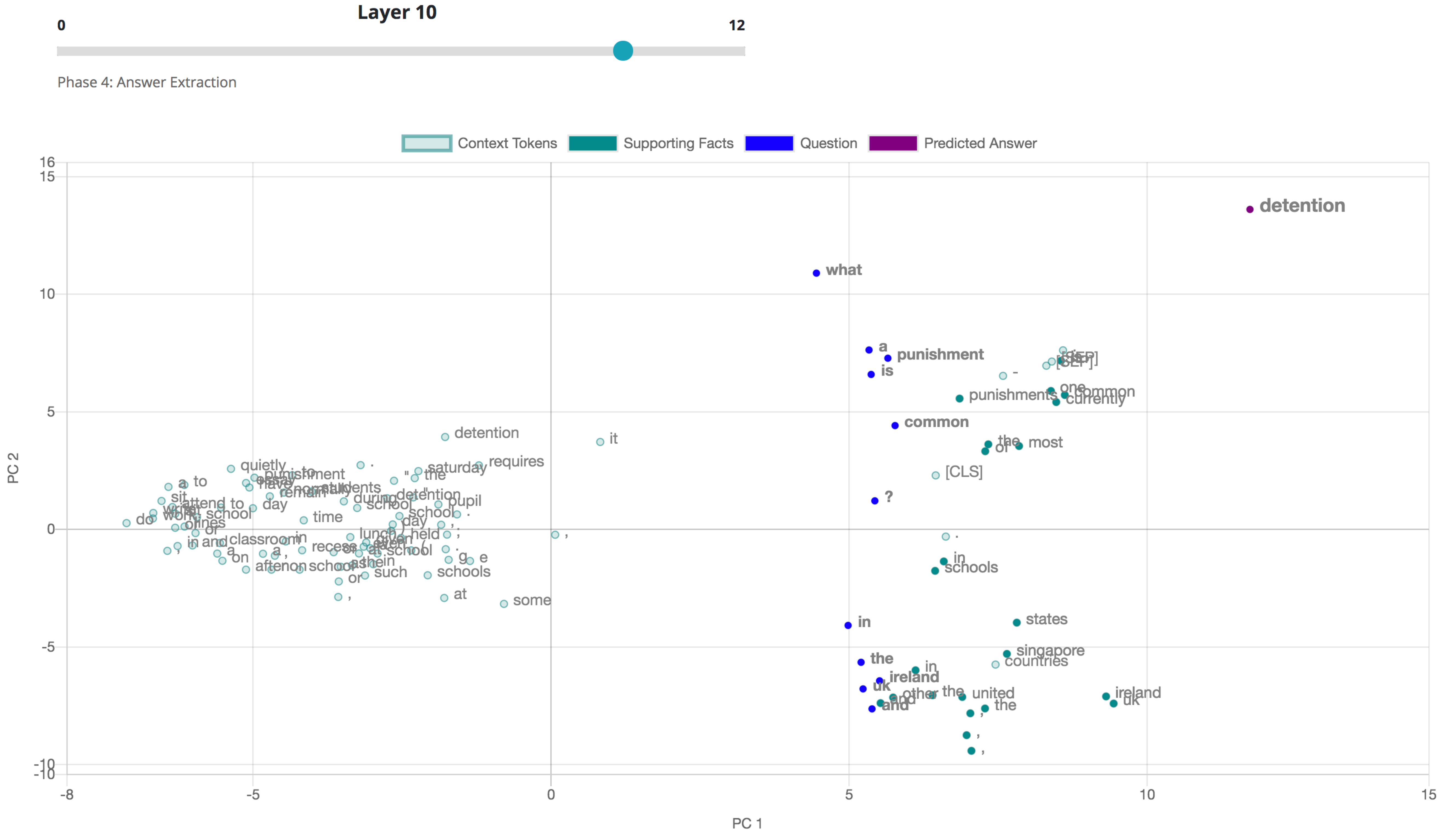}
  \caption{VisBERT's visualization interface: At the top, a slider to choose which of the 12 (BERT-base) or 24 (BERT-large) layers to visualize as well as an estimate, which of the four phases this layer belongs to. The legend below allows interactive filtering of different parts of the input. The main graph shows the contextual representations of the input tokens, each dot representing one token, color-coded by their affiliation. This example shows the SQuAD model in layer 10: One cluster contains irrelevant context tokens (left), one holds question tokens clustered together with supporting fact tokens (middle) and the predicted answer token (right) was separated from the rest.}
\label{fig:graph}
\end{figure*}

\section{Demonstration Outline}
The user interface of the browser-based VisBERT tool is shown in Figure \ref{fig:qa-box} and \ref{fig:graph}. We describe its application below.\\

\noindent\textbf{Included Question Answering Tasks}. We equip the tool with models for three well-known Question Answering tasks:
\begin{itemize}
    \item \textit{SQuAD}, the most popular recent QA dataset with 100,000 natural language questions. BERT-based models reach human performance on SQuAD.
    \item \textit{bAbI QA}, which is a collection of 20 different artificial toy tasks. The tasks contain simple patterns and are fully solved by recent models. However, they provide a useful testbed for clearly observing inference paths.
    \item \textit{HotpotQA} is the most difficult of the three tasks. Its 112,000 natural language questions come with long context texts and were especially created to require multi-hop inference.
\end{itemize}
We intentionally choose three diverse tasks in order to observe the influence of task design on BERT's hidden representations. For each task we provide a separate fine-tuned BERT model. We use a BERT-base model (12 layers) for SQuAD and bAbI and BERT-large (24 layers) for HotpotQA, because the base model does not produce adequate results on this more difficult task.  We also reduce the context size of HotpotQA samples that exceed BERT's 512 token limit. In addition to the included datasets, the tool can be easily extended to other Question Answering tasks.\\

\noindent\textbf{Sample Selection}. The tool includes a selection of samples from the test sets of each dataset. As the bAbI task comprises 20 different QA tasks, we choose exactly one sample per task and ignore the tasks that cannot be solved by span prediction (e.g. Positional Reasoning). In addition, the user is able to enter own examples. The requirement for these examples are to enter a question and a context document that contains the answer. The user can optionally enter the correct answer and the tool will automatically extract the sentence containing this answer as the supporting fact.\\

\noindent\textbf{Layer-Wise Visualization}. After selecting a sample, or entering one of their own, the user will get the prediction from the selected fine-tuned model. In addition to the predicted answer, a graph shows the token representations for a given layer. The representations are presented in 2D space after dimensionality reduction. Each point in the vector space represents one token. The tokens are color-coded into four categories: Question, supporting fact, context and predicted answer. This way the user can specifically analyse how the distances between certain tokens, e.g. question and supporting facts tokens, change. The user can also hide a group of tokens to only observe the remaining groups. By using the layer-slider on top of the graph, the user is able to go through all layers of the model and observe the changes within the token representations. This allows to inspect how the representations are influenced by the context and the underlying task over the layers of BERT.
\section{Observations}
VisBERT provides the possibility to explore the internal state of a model at each layer position. In the following we describe observations made from these internal states. These findings show how our tool can help to gain a better understanding of the inner workings of Transformer models.\\

\noindent\textbf{Phases of BERT's Inference}.
As shown in \cite{van2019does}, BERT models pass multiple phases while answering a question. VisBERT is able to demonstrate these phases in all three selected QA tasks despite their diversity. The tool indicates which phase is currently active, so that users can compare them with their own observations. We describe the phases briefly in the following:

\begin{enumerate}
    \item \textbf{Topical Clustering} In the first layers we see that tokens are clustered based on topical similarities, comparable to a static word embeddings like Word2Vec \cite{word2vec}.
    
    \item \textbf{Connecting Entities with Mentions and Attributes} Middle layers tend to cluster tokens based on their relation in the specific context. For example, we see multi-token entities clustered together because their tokens share one semantic meaning. One can also observe clusters of entities with their specific attributes.
    
    \item \textbf{Matching Questions with Supporting Facts} In the third quarter of BERT layers, we can see that the question tokens form clusters with the tokens of supporting facts. In multi-hop questions we even observe clusters for each \textit{hop} that the question contains.
    
    \item \textbf{Answer Extraction} In the last layers the answer tokens are separated from all other tokens. Earlier semantic clusters are dissolved. Based on the certainty of the decision, there might be other potential candidate tokens separated as well, with the furthest answer tokens being chosen as final prediction.
\end{enumerate}

\noindent\textbf{Adversarial Examples}. The system allows the input of new samples that do not belong to the preloaded test sets. On the one hand, this allows users to find out which QA model (SQuAD, HotpotQA or bAbI) fits a specific question type best and produces the right result. On the other hand, the tool can be used to explore how the models react to Adversarial Examples  \cite{adversarial-examples}. This way it is possible to discover potential deficits and biases within the model. For example, a user can add distracting facts to the context and check whether the model is still able to follow the same inference path. Effective methods for such adversarial examples on SQuAD are proposed by \cite{jia}. Our tool allows to not only observe resulting changes in the prediction, but also within the hidden states of a model.\\

\noindent\textbf{Failure States}. Decision legitimization is an important aspect of neural network explainability. If a network predicts an answer, it is useful to know why, in order to both improve the network and to understand its limits. VisBERT's visualizations show signs of wrong predictions not only in the last layers, even early phases can be helpful in analyzing errors. For example, in cases for which a wrong prediction has the same type as the ground truth answer, the problem is often that the wrong supporting fact was selected. This is clearly visible in layers of phase 3, where the question is matched with a wrong fact. For predictions that are completely wrong (not even of the same type as the answer) the phases often degenerate completely. This results in all layers looking either like a mostly homogeneous cloud of tokens or like they are stuck in phase 1, simply repeating the topical clustering with only slight reordering. Lastly, the network's general confidence can be estimated by looking at the clusters in each layer. For samples in which BERT is very confident, the clusters and phases are distinct. The lower the confidence, the more blurry and indistinct the clusters become.
\section{Conclusion}
VisBERT establishes a novel method to analyze the behavior of BERT models, in particular regarding the Question Answering task. Our method allows a fine-grained analysis of each of the BERT layers and depicts how each input token changes in each step. Additionally, VisBERT reveals four phases in BERT's transformations that are common to all of the datasets we examined and that mirror the traditional NLP pipeline, cf. \cite{nlp-pipeline}. We establish this behaviour on three diverse Question Answering datasets and make all three models available for users to make their own analyses on their own data, as well as the code to reproduce this visualization.\\

\noindent\textbf{Future Work}.
Our tool can easily be extended to other BERT models, fine-tuned on different QA datasets or even other NLP tasks entirely, and to other Transformer based models like GPT-2 \cite{gpt2}. Additionally it can be extended to include other dimensionality reduction methods like t-SNE or UMAP \cite{umap}.

Furthermore, we aim to explore the modularity demonstrated by the four phases we discovered in BERT's transformations. This modularity could be pushed even further, by fine-tuning different layers of BERT on different upstream tasks before training end-to-end on the final downstream task.

The ability to observe how wrong predictions are formed could be exploited for predicting a model's certainty even in early layers. Improvements on the model can be verified by observing changed behavior throughout its layers.

\begin{acks}
Our work is funded by the European Unions Horizon 2020 research and innovation programme under grant agreement 732328 (FashionBrain), by the German Federal Ministry of Education and Research (BMBF) under grant agreement 01UG1735BX (NOHATE) and by the German Federal Ministry of Economic Affairs and Energy (BMWi) under grant agreements 01MD19013D   (Smart-MD), 01MD19003E (PLASS) and 01MK2008D (Servicemeister).
\end{acks}

\bibliographystyle{ACM-Reference-Format}
\bibliography{www2020}

\end{document}